\documentclass[a4paper,conference]{IEEEtran}

\ifCLASSINFOpdf
\usepackage[pdftex]{graphicx}
\graphicspath{{../pdf/}{../jpeg/}}     
 \DeclareGraphicsExtensions{.pdf,.jpeg,.png}
 \else
 \usepackage[dvips]{graphicx}
 \graphicspath{{../eps/}}
 \DeclareGraphicsExtensions{.eps}
\fi

\ifCLASSINFOpdf
\else
\fi

\usepackage{graphicx}
\usepackage{subcaption}
\usepackage{threeparttable}

\begin{document}
%
\title{Transferring Rich Deep Features \\for Facial Beauty Prediction}

\author{\IEEEauthorblockN{\small Lu Xu}
\IEEEauthorblockA{\small College of Informatics\\
\small Huazhong Agricultural University\\
\small Wuhan, China \\
\small Email: xulu\_coi@webmail.hzau.edu.cn}
\and
\IEEEauthorblockN{\small Jinhai Xiang}
\IEEEauthorblockA{\small College of Informatics\\
\small Huazhong Agricultural University\\
\small Wuhan, China \\
\small Email: jimmy\_xiang@mail.hzau.edu.cn}
\and
\IEEEauthorblockN{\small Xiaohui Yuan}
\IEEEauthorblockA{\small Department of Computer Science and Engineering\\
\small University of North Texas\\
\small Denton, USA \\
\small Email: Xiaohui.Yuan@unt.edu}}


%


\maketitle

\begin{abstract}
Feature extraction plays a significant part in computer vision tasks.
In this paper, we propose a method which transfers rich deep features from
a pretrained model on face verification task and feeds the features into Bayesian
ridge regression algorithm for  facial beauty prediction.
We leverage the deep neural networks that extracts more abstract features from stacked
layers. Through simple but effective feature fusion strategy, our method achieves
improved or comparable performance on SCUT-FBP dataset and ECCV HotOrNot dataset.
Our experiments demonstrate the effectiveness of the proposed method and clarify the inner
interpretability of facial beauty perception.

\end{abstract}


%
\IEEEpeerreviewmaketitle

\section{Introduction}

  Facial beauty analysis~\cite{Perrett1994Facial} has been widely used in many
  fields such as facial image beautification APPs (e.g., MeiTu and Facetune),
  plastic surgery, and face-based pose analysis~\cite{Liu17}.
  In the mobile computing era, billions of images per day are acquired and uploaded
  to social networks and online platform, leading to the demand for
  better image processing and analyzing technology. Recently, thanks to the
  big data and high-performance computational hardware,
  computational and data-driven approaches have been proposed for
  solving these questions such as face recognition, facial
  expression recognition, facial beauty analysis and etc.

  The existing methods resort to machine learning and computer vision techniques
  to analyze facial beauty and achieve promising results \cite{zhang2016computer}.
  The methods often include image feature descriptors (such as HOG, SIFT, LBP, etc)
  and supervised machine learning predictors (such as SVM, KNN, DNN, LR, etc).

  In order to explore the best facial beauty prediction approach that precisely
  maps high-level features into face beauty ratings, we propose a method that
  combines transfer learning and Bayesian regression. The method achieves the
  improved or comparable performance on SCUT-FBP dataset~\cite{xie2015scut} and
  ECCV HotOrNot dataset~\cite{gray2010predicting}.

  The main contributions of this paper are  as follows:
\begin{itemize}
  \item
    We apply transfer learning to our facial beauty prediction problems for
    feature extraction. Experimental results show that the transferred deep
    features can attain more impressive performance compared with the
    traditional image feature descriptors such as HOG, LBP and gray
    value features.
  \item
    We make a detailed analysis about deep features based on knowledge adaptation.
    Additionally, we perform an effective feature fusion strategy to build more
    informative facial features in our facial beauty prediction task.
  \item
    Studies found that the neural networks are lack of satisfactory
    interpretation. We make ablative studies by visualizing the face feature and
    reveal the elements that influence facial beauty perception.

\end{itemize}

  The rest of this paper is organized as follows. Section
  \uppercase\expandafter{\romannumeral2} reviews the related works of facial descriptor and learning methods.
  Section \uppercase\expandafter{\romannumeral3} describes our proposed method
  in details, which include deep feature extraction and Bayesian ridge regression. Experimental
  results and comparisons are presented in Section
  \uppercase\expandafter{\romannumeral4} and Section
  \uppercase\expandafter{\romannumeral5} concludes this paper with a summary and future work.


\section{RELATED WORK}

  \subsection{Facial Descriptors and Machine Learning Predictors}
    Many researchers focus on developing new machine learning algorithms to
    achieve better classification or regression performance, while others focus
    on designing better facial feature descriptors.
    Zhang et al. \cite{chen2016data} combine several low-level face
    representations and high-level features to form a feature vector and
    perform feature selection to optimize the feature set.
    Eisenthal et al. \cite{eisenthal2006facial} use a vector of gray values
    created by concatenating the rows or columns of an image.
    Huang et al. \cite{huang2012learning} propose a method to learn hierarchical
    representations of convolutional deep belief networks.
    Xie et al. \cite{xie2015scut} resort to deep learning to train a predictor
    and achieve state-of-the-art performance.
    Amit et al. \cite{kagian2007humanlike} use numerous facial features that
    describe facial geometry, color and texture to predict facial attractiveness.
    Lu et al. \cite{Lu2016A} detect face landmarks with ASM and then extract
    facial features based on Blocked-LBP which achieved the Pearson Correlation
    at 0.874 on 400 high-quality female face images.
    Zhang et al. \cite{zhang2011quantitative} compute geometric distances
    between feature points and ratio vectors composed of geometric distances,
    and then treat them as features for machine learning algorithm.
    For the lack of abundant labeled images, it always takes lots of
    time to fine-tune the deep neural networks architecture and parameters to
    achieve a comparative result and avoid overfitting problems as well.

    In addition, some research works towards developing or improving new machine
    learning algorithms.
    Eisenthal et al. \cite{eisenthal2006facial} employ KNN and SVM as classifiers
    to rate faces belongs to different levels.
    Gan et al. \cite{gan2014deep} use deep self-taught learning to obtain
    hierarchical representations and learn the concept of facial beauty.
    Xu et al. \cite{xu2015new} propose a method which constructs a convolutional
    neural network (CNN) for facial beauty prediction using
    a new deep cascaded fine tuning scheme with various face inputting channels.
    Wang et al. \cite{wang2014attractive} use deep auto encoders to extract
    features and take a low-rank fusion method to integrate scores, and their
    method achieves promising results.
    Xu et al. \cite{Xu2017Facial} propose  ``psychologically
    inspired CNN (PI-CNN)'' for automatically facial beauty prediction.

  \subsection{Deep CNN and Transfer Learning}
    Deep learning allows computational models that are composed of
    processing layers to learn representations of data with multiple levels of
    abstraction \cite{lecun2015deep}. CNN is a type of neural networks which is
    designed to process data that come in form of multiple arrays.
    Deep learning has been used as a dramatically powerful tool in computer
    vision tasks such as image recognition
    \cite{krizhevsky2012imagenet,simonyan2014very,szegedy2015going,he2016deep}.
    The features are automatically extracted via stacked layers.
    Neural networks are trained through
    back-propagation algorithm to minimize the cost function.

    Deep convolutional neural networks show more extraordinary capacity in
    feature extraction than traditional hand-crafted descriptors. However, we may
    need to design different networks architectures and train the deep neural
    networks almost from scratch to satisfy our task, which takes much
    computational burden.
    Transfer learning allows us to fine-tune the higher layers based on a
    pretrained model, or even just treat the pretrained model as a feature extractor.

    Yosinski et al. \cite{yosinski2014transferable} show that initializing
    a network with transferred features from almost any number of layers can
    produce an improvement to the generalization even after fine-tuning to
    the target domain dataset.
    Yoshua Bengio et al. \cite{bengio2012deep} explore why unsupervised
    pre-training of representations can be useful, and how it can be exploited
    in the transfer learning scenario.
    Donahue et al. \cite{donahue2014decaf} show that the features extracted by
    deep convolutional neural networks pretrained on ImageNet can achieve much
    better performance than many algorithms on lots of classification tasks,
    which illustrates the great generality and transferability of deep
    convolutional neural networks.

\section{Method}

 \subsection{VGG Network}
\begin{figure}[!htb]
      \centering
      \includegraphics[scale=0.2]{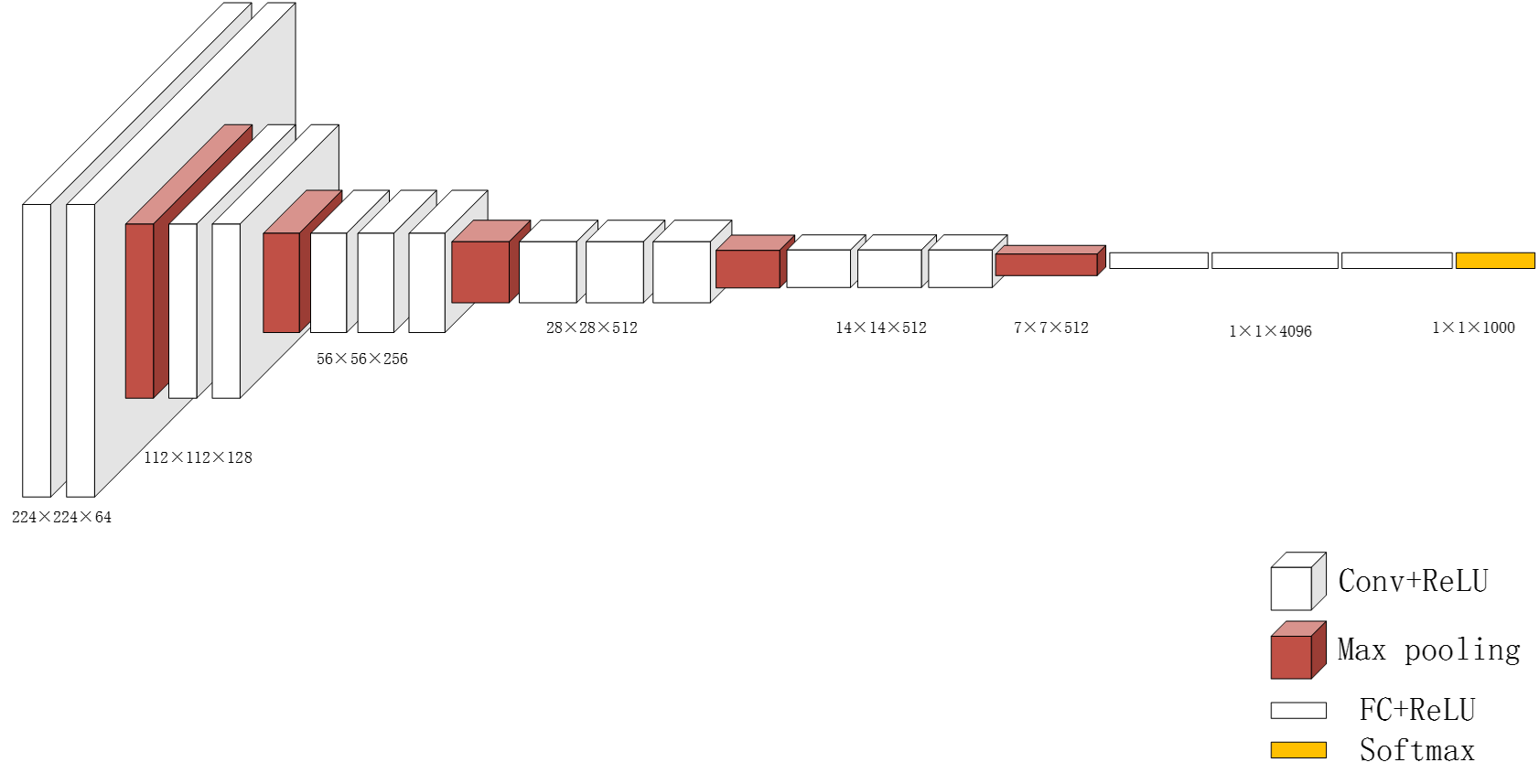}
      \caption{Networks architecture. we adopt VGG16 in our feature extraction
      procedure, which is composed of multiple small convolutional filters to
      extract more informative features compared with bigger filters used in
      AlexNet~\cite{krizhevsky2012imagenet} for ImageNet recognition task.}
      \label{vgg16}
    \end{figure}

    We include a brief review of VGG, which is employed by our proposed method.
    VGG~\cite{simonyan2014very} consists of 16-19 weight layers and very small
    ($3\times3$) convolution filters as well.
    Fig.~\ref{vgg16} shows the overall architecture of the \emph{VGG16} networks.
    Though VGG networks architecture is simple, it is widely used in many
    computer vision tasks. In our experiments, we take a VGG face model which is
    pretrained on a face verification task \cite{parkhi2015deep}.
    Although the original task is absolutely different from our facial beauty
    prediction task, it shows dramatically impressive performance.
    We believe the main reason for this issue can be attributed to the
    extraordinary feature representation power of deep CNNs.

  \subsection{Deep Feature Extraction}
    Several research works \cite{bengio2012deep,lecun2015deep} show
    that the deep convolutional neural networks can learn increasingly powerful
    representations as the feature hierarchy becomes deeper. However,
    due to the limited labeled face images, if we train a deep convolutional
    neural network directly, we may suffer from severe overfitting problems.
    Recently, transfer learning has aroused much attention~\cite{Yuan18}, which
    enables us to fine-tune from a pretrained model or just treat the learned
    neural network as a feature extractor to satisfy our tasks
    \cite{yosinski2014transferable}.

    We extract facial features with VGG face model \cite{parkhi2015deep}
    pretrained on face verification task.
    Despite their target task is different from our facial
    beauty prediction task, the feature can achieve remarkable performance, which
    indicates extraordinary feature representation power of CNNs to some degree.
    Researches \cite{yosinski2014transferable} show that the features in lower
    layers contain more detailed information while features in higher layers
    represent more semantic meaning.
    Our method concatenates on
    both relatively low layer's features and relatively high layer's features as our
    facial representation.
    We also use HOG, grayscale and LBP features in our experiments for
    comparison to evaluate the feature extraction capacity of deep CNNs.

  \subsection{Bayesian Ridge Regression}
    We feed the concatenated feature vectors into Bayesian ridge regressor.
    Bayesian ridge regression includes regularization during estimation procedure:
    the regularization item is not embedded with cost function directly,
    but tuned to your data distribution. The $L_2$
    regularization used in Bayesian ridge regression is equal to maximizing
    a posterior estimation of the parameters $w$
    with precision $\lambda^{-1}$ under a Gaussian prior.

    The output $y$ is assumed to be Gaussian distributed around $Xw$
    in order to form a fully probabilistic model:

    \begin{equation}
      P\left(y\middle|X,w,\alpha\right)=\mathcal N\left(y\middle|Xw,\alpha\right)
    \end{equation}

    Bayesian ridge regressor evaluates a probabilistic model of the regression
    problem. The prior for the parameter $w$ is decided by a spherical Gaussian:
    \begin{equation}
      P\left(w\middle|\lambda\right)=\mathcal N\left(w\middle|0,\lambda^{-1}I_p\right)
    \end{equation}

    The priors over $\alpha$ and $\lambda$ are chosen to be Gamma distributions,
    the conjugate prior for the precision of the Gaussian.

    The parameters $w$, $\alpha$ and $\lambda$ are estimated jointly during the
    fit procedure. The remaining hyperparameters are the parameters of the
    Gamma priors over $\alpha$ and $\lambda$.
    All the parameters are tuned by maximizing the marginal log likelihood.

\section{Experiments}
  We implement our method with TensorFlow~\cite{abadi2016tensorflow}
  and Scikit-Learn~\cite{pedregosa2011scikit} on an Ubuntu server with NVIDIA
  Tesla K80 GPU and Intel Xeon CPU.

  \subsection{SCUT-FBP Dataset}
    The SCUT-FBP dataset~\cite{xie2015scut} contains images of 500 Asian females.
    Each image is scored by 10 raters, the main task is to build a computational
    model to predict the average score of the human portrait image.

\begin{figure}[htb]
      \centering
      \includegraphics[scale=0.25]{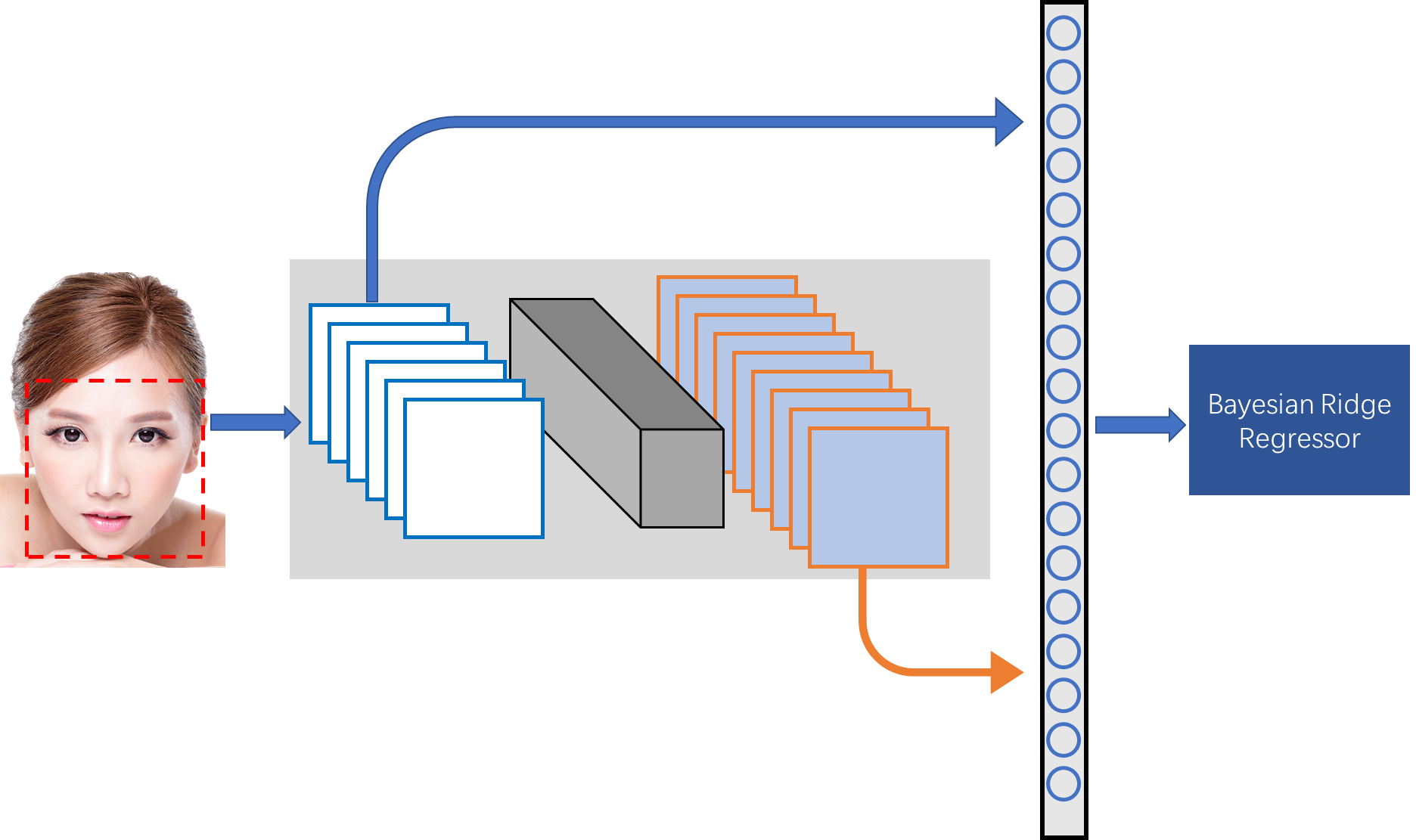}
      \caption{Pipeline of our proposed method. The face is detected and
        then fed into CNNs, we concatenate $conv4\_1$ and $conv5\_1$ layers'
        feature maps, and flatten them into feature vectors for the input of Bayesian
        ridge regression.}
      \label{pipeline}
    \end{figure}

    Since the images in SCUT-FBP~\cite{xie2015scut} are not in same size,
    deep CNNs can only support fixed squared data as input. We conduct three
    methods named ``Crop'', ``Warp'' and ``Padding'' to get squared images
    respectively.
    In ``Crop'' setting, we detect face provided by \cite{King2009Dlib}
    and crop the face region, then we resize it to $224\times224$.
    In ``Warp'' setting, we just warp the image forcely to form a
    $224\times224$ image.
    In ``Padding'' setting, we resize the longer side
    to 224 and zero-pad the shorter side to form a $224\times224$ image
    (See Fig.~\ref{warp_and_padding}). We also normalize the input image
    by substracting the mean and dividing the standard variance of the pixels.
    Furthermore, we manually crop the central region of the image and treat it
    as the input for our neural networks in case of failed face detection.
    In SCUT-FBP dataset, we concatenates the \emph{conv5\_1} and \emph{conv4\_1}
    layer's features. The pipeline is shown in Fig.~\ref{pipeline}:

    \begin{figure}
      \centering
      \begin{subfigure}[b]{0.1\textwidth}
          \includegraphics[width=\textwidth]{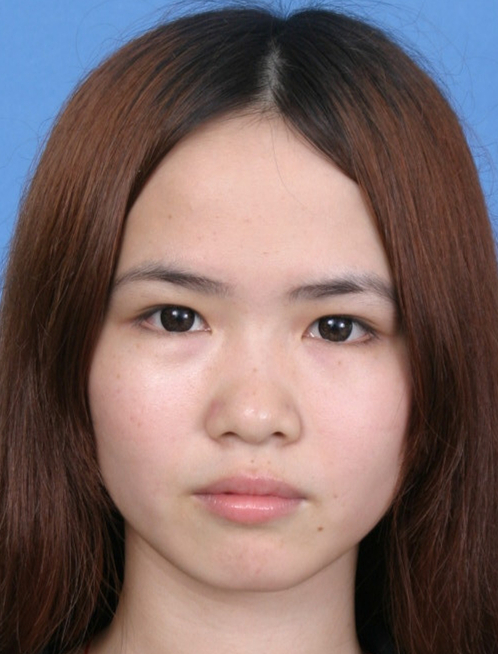}
          \caption{Original}
          \label{fig:original}
      \end{subfigure}
      ~
      \begin{subfigure}[b]{0.1\textwidth}
          \includegraphics[width=\textwidth]{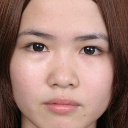}
          \caption{Crop}
          \label{fig:crop}
      \end{subfigure}
      ~
      \begin{subfigure}[b]{0.1\textwidth}
          \includegraphics[width=\textwidth]{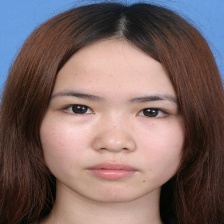}
          \caption{Warp}
          \label{fig:warp}
      \end{subfigure}
      ~
      \begin{subfigure}[b]{0.1\textwidth}
          \includegraphics[width=\textwidth]{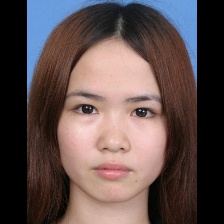}
          \caption{Padding}
          \label{fig:padding}
      \end{subfigure}
      \caption{Different settings to form a squared image:
        (a) Original image in SCUT-FBP.
        (b) Cropped image.
        (c) Image warp.
        (d) Image padding.
        We conduct these experiments to see whether the facial beauty perception
        is correlated with non-facial elements such as haircut, wearing, posture
        and etc.
      }
      \label{warp_and_padding}
    \end{figure}

  \subsection{Performance Evaluation}
    In our experiment, we use Pearson Correlation (PC),
    Mean Absolute Error (MAE) and Root Mean Squared Error (RMSE) as
    the criteria for evaluating our method.

    \begin{equation}
      RMSE(X,h)=\sqrt{\frac{1}{m}\sum_{i=1}^m\big(h(x^{(i)})-y^{(i)})^2}
    \end{equation}

    \begin{equation}
      MAE=\frac{1}{m} \sum_{i=1}^m|h(x^{(i)})-y^{(i)}|
    \end{equation}

    \begin{equation}
      PC=\frac{1}{n-1}\sum_{i=1}^n
      (\frac{h(x_i)-\bar{h(x)}}{s_{h(x)}})(\frac{y_i-\bar{y}}{s_y})
    \end{equation}
    where $\emph{m}$ denotes the number of images, $x^{(i)}$ denotes the
    input feature vector of image $i$, $h(\bullet)$ denotes the learning algorithm,
    $y^{(i)}$ denotes the groundtruth attractiveness score of image $i$.

    MAE and RMSE measure the fit quality of the learning algorithms,
    the performance is better if the value is closer to zero.
    PC measures the linear correlation between $h(x^{(i)})$ and $y^{(i)}$.
    Its value lays between 1 and -1, where 1 means absolutely positive linear
    correlation, 0 means no linear correlation, and -1 means absolutely negative
    linear correlation.


    In order to make the prediction more reliable and reproducable,
    we follow the provision denoted in \cite{xie2015scut} for fair comparison.
    We randomly select 400 images as training set and the rest 100 images as
    test set. Finally, we average the 5 experimental results as the final
    performance to remove sample variances. The results are shown in
    TABLE \ref{avg performance}.
    \begin{table}[!htb]
      \caption{AVERAGE PERFORMANCE}
      \label{avg performance}
      \begin{center}
        \begin{tabular}{ l | c c c }
        \hline
          & \textbf{MAE} & \textbf{RMSE} & \textbf{PC} \\ \hline\hline
        1 & 0.2569 & 0.3418 & 0.8735 \\
        2 & 0.2594 & 0.3470 & 0.8299 \\
        3 & 0.2479 & 0.3117 & 0.8929 \\
        4 & 0.2651 & 0.3473 & 0.8562 \\
        5 & 0.2680 & 0.3508 & 0.8323 \\ \hline
        AVG & \textbf{0.2595} & \textbf{0.3397} & \textbf{0.8570}\\
        \hline
        \end{tabular}
        \begin{tablenotes}
          \footnotesize
          Performance on SCUT-FBP~\cite{xie2015scut} of 5 rounds. We average
          the 5 results to remove examples variances as denoted in \cite{xie2015scut}.
        \end{tablenotes}
      \end{center}
    \end{table}

    TABLE \ref{performance comparison} shows performance comparison with other
    methods. The best performance is marked with bold font and the second best
    is highlighted with an underline.
    Our method ranks the second place on SCUT-FBP~\cite{xie2015scut} dataset.
   \begin{table}[!hbt]
      \caption{PREDICTOR PERFORMANCE COMPARISON}
      \label{performance comparison}
      \begin{center}
        \begin{tabular}{l|ccr}
          \hline
          \textbf{Method} & \textbf{RMSE} & \textbf{MAE} & \textbf{PC} \\ \hline\hline
          KeyPointGabor+PCA+SVR & 0.5606 & 0.5541 & 0.5490 \\
          KeyPointGabor+PCA+Gaussian Reg & 0.6152 & 0.4724 & 0.4591 \\
          UniSampleGabor+PCA+SVR & 0.5452 & 0.4230 & 0.5847 \\
          UniSampleGabor+PCA+Gaussian Reg & 0.5164 & 0.3969 & 0.6347 \\
          Combined Features+SVR~\cite{xie2015scut} & \underline{0.5120} & 0.3961 & 0.6433 \\
          Combined Features+Gaussian Reg~\cite{xie2015scut} & 0.5149 & \underline{0.3931} & 0.6482 \\
          CNN-based \cite{xie2015scut} & - & - & 0.8187 \\
          PI-CNN \cite{Xu2017Facial} & - & - & {\textbf{0.87}} \\
          Ours & {\textbf{0.2595}} & {\textbf{0.3397}} & {\underline{0.8570}} \\ \hline
        \end{tabular}
        \begin{tablenotes}
          \footnotesize
          Performance comparison with other methods, our method ranks the second
          place on PC and first place on RMSE and MAE, respectively.
          The best and second results are emphasized in bold and underline respectively.
          Since RMSE and MAE of CNN-based methods proposed in
          \cite{xie2015scut} and \cite{Xu2017Facial} are not given and are hence
          denoted with ``-".
        \end{tablenotes}
      \end{center}
    \end{table}

  \subsection{Ablation Analysis}
    It is almost a common sense in machine learning practice is that ``feature
    matters".
    To illustrate the feature extraction capability by deep learning,
    we conduct experiments based on different features including HOG, LBP,
    gray image and transferred deep features for performance comparison
    and visualization:

    \begin{figure}
      \centering
      \begin{subfigure}[b]{0.08\textwidth}
          \includegraphics[scale=0.5]{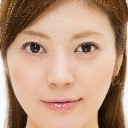}
          \label{fig:origin_1}
      \end{subfigure}
      ~
      \begin{subfigure}[b]{0.08\textwidth}
          \includegraphics[scale=0.5]{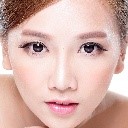}
          \label{fig:origin_2}
      \end{subfigure}
      ~
      \begin{subfigure}[b]{0.08\textwidth}
          \includegraphics[scale=0.5]{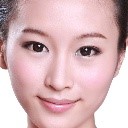}
          \label{fig:origin_3}
      \end{subfigure}
      \\
      \vskip 0.1em
      \begin{subfigure}[b]{0.08\textwidth}
          \includegraphics[scale=0.92]{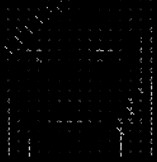}
          \label{fig:hog_1}
      \end{subfigure}
      ~
      \begin{subfigure}[b]{0.08\textwidth}
          \includegraphics[scale=0.92]{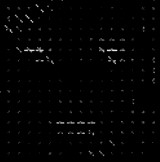}
          \label{fig:hog_2}
      \end{subfigure}
      ~
      \begin{subfigure}[b]{0.08\textwidth}
          \includegraphics[scale=0.92]{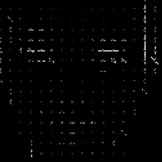}
          \label{fig:hog_3}
      \end{subfigure}
      \\
      \vskip 0.1em
      \begin{subfigure}[b]{0.08\textwidth}
          \includegraphics[scale=0.5]{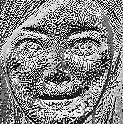}
          \label{fig:lbp_1}
      \end{subfigure}
      ~
      \begin{subfigure}[b]{0.08\textwidth}
          \includegraphics[scale=0.5]{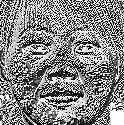}
          \label{fig:lbp_2}
      \end{subfigure}
      ~
      \begin{subfigure}[b]{0.08\textwidth}
          \includegraphics[scale=0.5]{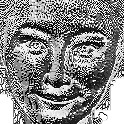}
          \label{fig:lbp_3}
      \end{subfigure}
      \caption{Visualization of the original portrait images and their
      corresponding visualized features described by different feature
      extractors:
      (top) Original portrait images.
      (middle) Visualization of HOG descriptors.
      (down) Visualization of LBP descriptors.
      }
      \label{fig:portrait images}
  \end{figure}

  \begin{itemize}
    \item \textbf{Raw Grayscale}: we convert the RGB facial images into their
          corresponding gray scale ones, and the flattened pixel gray scale
          value is used as the feature.
    \item \textbf{HOG}: HOG is an image feature descriptor which is widely used
          in computer vision and image processing for object detection tasks.
          Details can be found in \cite{dalal2005histograms}.
    \item \textbf{LBP}: LBP is a type of feature descriptor which especially
          cares more about texture details, and is widely used in many machine
          vision tasks.
  \end{itemize}

  \begin{table}
    \caption{PERFORMANCE COMPARISON BETWEEN DIFFERENT FEATURES}
    \label{different features performance comparison}
    \begin{center}
      \begin{tabular}{l|ccr}
        \hline
        \textbf{Feature} & \textbf{RMSE} & \textbf{MAE} & \textbf{PC} \\ \hline\hline
        TransCNN & {\textbf{0.2595}} & {\textbf{0.3397}} & {\textbf{0.8570}} \\
        HOG & 0.3308 & 0.4394 & 0.6216 \\
        LBP & 0.3987 & 0.4800  & 0.5631  \\
        Gray Scale & 0.4008 & 0.4889 & 0.5149 \\ \hline
      \end{tabular}
      \begin{tablenotes}
        \footnotesize
        Performance comparison with other feature descriptors on Bayesian ridge
        regression. Our transferred deep features outperforms other descriptors
        with a large margin. The best results are given in bold font.
      \end{tablenotes}
    \end{center}
  \end{table}

\begin{figure}[!htb]
    \centering
    \includegraphics[scale=0.5]{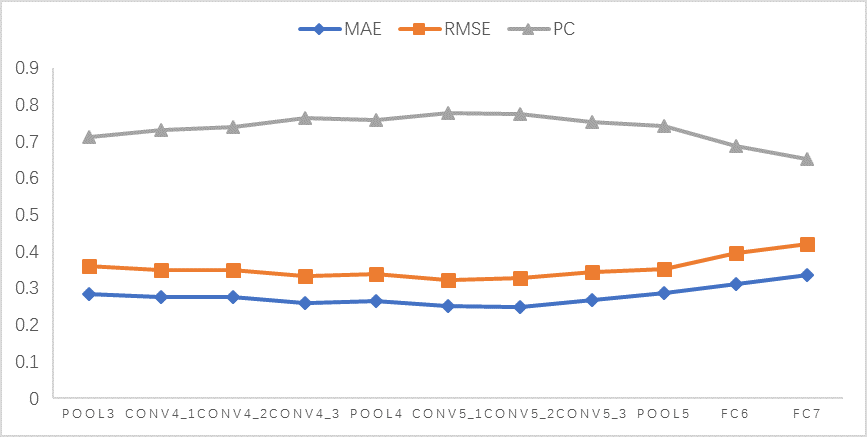}
    \caption{Performance comparison between different layers: the performance
    gets better as layer goes deeper, which means the deep CNN extracts more
    discriminative features. It decreases sharply after max pooling operation,
    which may be attributed as heavy spatial information loss.}
    \label{deep feature comparison}
  \end{figure}

  In addition, we compare the feature performance from different layers to find
  which layer produces the most discriminative features (See Fig.~\ref{deep feature comparison}).

  Moreover, among three preprocessing methods (Crop, Warp, and Padding),
  Crop achieves the best performance on SCUT-FBP, which indicates that facial
  region plays a more significant part in beauty perception, while background may
  act as noise in our facial beauty prediction task on SCUT-FBP dataset
  (See TABLE.~\ref{pc of diffrent preprocessing}).

  Fig. \ref{deep feature comparison} depicts that as layer goes deeper
  the performance gets better, and reaches the best at \emph{conv5\_1}.
  While when feature maps are flattened into vectors, we see a sharp drop
  in performance, which may be attributed as the heavy spatial information loss.

  \begin{table}[!htb]
    \caption{PEARSON CORRELATION OF DIFFERENT PREPROCESSING METHODS}
    \label{pc of diffrent preprocessing}
    \begin{center}
      \begin{tabular}{ l | c c c }
      \hline
        & \textbf{Crop} & \textbf{Warp} & \textbf{Padding} \\ \hline\hline
      PC & \textbf{0.8570} & 0.7376 & 0.8255\\
      \hline
      \end{tabular}
      \begin{tablenotes}
        \footnotesize
        Performance of diffrent preprocessing methods
        (``Crop'', ``Warp'', and ``Padding'') on SCUT-FBP.
        ``Crop'' achieves the best.
      \end{tablenotes}
    \end{center}
  \end{table}

  \subsection{ECCV HotOrNot Dataset}
\begin{figure}
    \centering
      \begin{subfigure}[b]{0.07\textwidth}
        \includegraphics[scale=0.35]{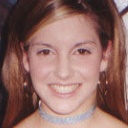}
        \caption{}
      \end{subfigure}
      ~
      \begin{subfigure}[b]{0.07\textwidth}
        \includegraphics[scale=0.35]{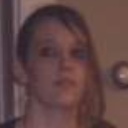}
        \caption{}
      \end{subfigure}
      ~
      \begin{subfigure}[b]{0.07\textwidth}
        \includegraphics[scale=0.35]{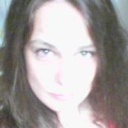}
        \caption{}
      \end{subfigure}
      ~
      \begin{subfigure}[b]{0.07\textwidth}
        \includegraphics[scale=0.35]{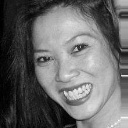}
        \caption{}
      \end{subfigure}
      \\
      \begin{subfigure}[b]{0.07\textwidth}
        \includegraphics[scale=0.35]{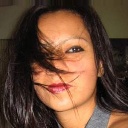}
        \caption{}
      \end{subfigure}
      ~
      \begin{subfigure}[b]{0.07\textwidth}
        \includegraphics[scale=0.35]{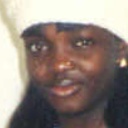}
        \caption{}
      \end{subfigure}
      ~
      \begin{subfigure}[b]{0.07\textwidth}
        \includegraphics[scale=0.35]{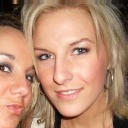}
        \caption{}
      \end{subfigure}
      ~
      \begin{subfigure}[b]{0.07\textwidth}
        \includegraphics[scale=0.35]{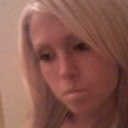}
        \caption{}
      \end{subfigure}
      \caption{ECCV HotOrNot face samples: this dataset is more challenging due
            to low resolution (b), illumination problem (c), gray version (d),
            occlusion (e), different race (f), cluttered background (g) and
            unaligned posture (h). Aligned samples are like (a).}
      \label{fig:eccv dataset}
    \end{figure}

    ECCV HotOrNot dataset~\cite{gray2010predicting} contains 2056 faces which
    are collected from the Internet. Each face is labeled with a score, and
    the dataset has already been split into 5 training and test datasets.
    Unlike SCUT-FBP dataset~\cite{xie2015scut}, the faces in ECCV HotOrNot
    dataset~\cite{gray2010predicting} are more challenging because of the variant
    postures, cluttered background, illumination, low resolution and unaligned
    faces problems, which make the facial beauty prediction more difficult
    (See Fig.~\ref{fig:eccv dataset}).

    ECCV HotOrNot dataset uses \emph{Pearson Correlation (PC)} for performance
    metric. We also list \emph{MAE} and \emph{RMSE} for more detailed comparison.

  \subsection{Ablation Study}
    We concatenate $conv5\_2$ and $conv5\_3$ layers' feature maps and flatten them
    to form more informative features. The concatenated features are then fed into
    Bayesian ridge regression algorithm \cite{mackay1992bayesian}.

    We implement two means to evaluate the impact of preprocessing techniques.
    In solution A, we run face detector~\cite{King2009Dlib} to detect
    68 facial landmarks and the facial region. For grayscale images, we replicate
    the gray pixel value twice to form an RGB channels image.
    Then we calculate the inclination angle to the horizontal line with two eyes
    coordinates, which is denoted as $\theta$.
    If $|\theta|>0$, we rotate the face around the central point by $\theta$ degree
    and crop the facial region. The mean pixel value is subtracted from the cropped image,
    which is normalized by its standard deviation.
    Solution B includes mean subtraction and standard error division on the
    original images. No additional preprocessing is taken.

    \begin{figure}[!htb]
      \centering
      \begin{subfigure}[b]{0.06\textwidth}
        \includegraphics[scale=0.3]{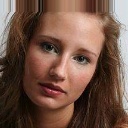}
      \end{subfigure}
      ~
      \begin{subfigure}[b]{0.06\textwidth}
        \includegraphics[scale=0.3]{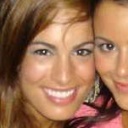}
      \end{subfigure}
      ~
      \begin{subfigure}[b]{0.06\textwidth}
        \includegraphics[scale=0.3]{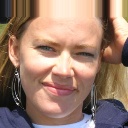}
      \end{subfigure}
      ~
      \begin{subfigure}[b]{0.06\textwidth}
        \includegraphics[scale=0.3]{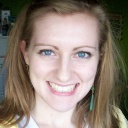}
      \end{subfigure}
      \\
      \vskip 0.3em
      \begin{subfigure}[b]{0.06\textwidth}
        \includegraphics[scale=0.3]{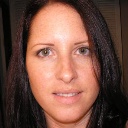}
      \end{subfigure}
      ~
      \begin{subfigure}[b]{0.06\textwidth}
        \includegraphics[scale=0.3]{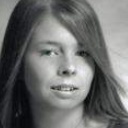}
      \end{subfigure}
      ~
      \begin{subfigure}[b]{0.06\textwidth}
        \includegraphics[scale=0.3]{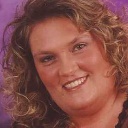}
      \end{subfigure}
      ~
      \begin{subfigure}[b]{0.06\textwidth}
        \includegraphics[scale=0.3]{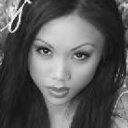}
      \end{subfigure}
      \caption{$\epsilon\geq2.75$, which means these samples are not well
      predicted by our algorithm.}
      \label{fig:bad prediction}
    \end{figure}

    \begin{figure}[!htb]
      \centering
      \begin{subfigure}[b]{0.06\textwidth}
        \includegraphics[scale=0.3]{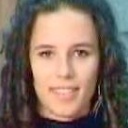}
      \end{subfigure}
      ~
      \begin{subfigure}[b]{0.06\textwidth}
        \includegraphics[scale=0.3]{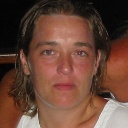}
      \end{subfigure}
      ~
      \begin{subfigure}[b]{0.06\textwidth}
        \includegraphics[scale=0.3]{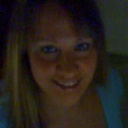}
      \end{subfigure}
      ~
      \begin{subfigure}[b]{0.06\textwidth}
        \includegraphics[scale=0.3]{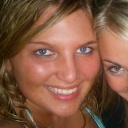}
      \end{subfigure}
      \\
      \vskip 0.3em
      \begin{subfigure}[b]{0.06\textwidth}
        \includegraphics[scale=0.3]{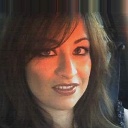}
      \end{subfigure}
      ~
      \begin{subfigure}[b]{0.06\textwidth}
        \includegraphics[scale=0.3]{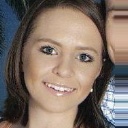}
      \end{subfigure}
      ~
      \begin{subfigure}[b]{0.06\textwidth}
        \includegraphics[scale=0.3]{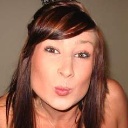}
      \end{subfigure}
      ~
      \begin{subfigure}[b]{0.06\textwidth}
        \includegraphics[scale=0.3]{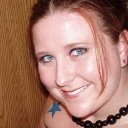}
      \end{subfigure}
      \caption{$\epsilon\leq0.02$, which means these samples are well fitted
      by our algorithm.}
      \label{fig:good prediction}
    \end{figure}

    \begin{table}[!htb]
      \caption{PERFORMANCE ON ECCV HOTORNOT DATASET}
      \label{eccv performance}
      \begin{center}
        \begin{tabular}{c|ccc|ccc}
          \hline
          \multicolumn{1}{l|}{} &
          \multicolumn{3}{|c|}{solution A} &
          \multicolumn{3}{|c}{solution B} \\
          \cline{2-7}
          \textbf{Dataset} & \textbf{RMSE} & \textbf{MAE} & \textbf{PC} &
          \textbf{RMSE} & \textbf{MAE} & \textbf{PC} \\ \hline\hline
          1 & 0.9417 & 1.1948 & 0.3970 & 0.9140 & 1.1493 & 0.4656 \\
          2 & 0.9755 & 1.2406 & 0.4022 & 0.9210 & 1.1562 & 0.4728 \\
          3 & 0.9293 & 1.1810 & 0.3840 & 0.8989 & 1.1258 & 0.4694 \\
          4 & 0.9469 & 1.1788 & 0.3898 & 0.8736 & 1.1087 & 0.4775 \\
          5 & 0.9394 & 1.1856 & 0.3862 & 0.9104 & 1.1316 & 0.4541 \\
          Average & \textbf{0.9466} & \textbf{1.1962} & \textbf{0.3918}
          &\textbf{0.9036} & \textbf{1.1343} & \textbf{0.4679} \\ \hline
        \end{tabular}
        \begin{tablenotes}
          \footnotesize
          The ECCV HotOrNot dataset \cite{gray2010predicting} has been divided
          into 5 parts which contain training set and test set respectively.
          We compare the performance of solution A and solution B.
          Much to our surprise, solution B achieves better results with a large
          margin. It may be explained by the extra non-facial information such
          as hairstyle, wearing, posture, etc.
        \end{tablenotes}
      \end{center}
    \end{table}

    We find that \emph{solution B} achieves much better performance
    than \emph{solution A}, the results can be found in TABLE \ref{eccv performance}.
    We believe the main reason is that the annotators may also take extra information
    such as haircut, posture, and clothing into consideration while labeling these
    facial beauty scores, instead of just measuring face region.

    Additionally, we define $\epsilon=|y^i-\hat{y}^i|$, which describes the error
    between the predicted facial beauty score ($\hat{y}^i$) and the ground truth
    beauty score ($y^i$). If $\epsilon\geq\tau_1$, we believe there is a relatively
    severe bias among the predicted values and ground truth scores. If
    $\epsilon\leq\tau_2$, we believe our algorithm fits these samples perfectly.

    In this part, we set $\tau_1=2.75$ and $\tau_2=0.02$ for detailed analysis
    (See Fig. \ref{fig:bad prediction} and Fig. \ref{fig:good prediction}).
    We believe the performance could be greatly improved through face alignment
    techniques. Besides, posture and facial expression may also contribute to
    beauty perception because our algorithm fails to capture these samples with
    variant postures.

    Table~\ref{tb:PC} compares the Pearson Correlation of our proposed method
    with five state-of-the-art methods. Our method outperforms other methods and
    achieves the best performance on ECCV HorOrNot dataset without face alignment.

    \begin{table}[!htb]
      \caption{PEARSON CORRELATION OF HOTORNOT DATASET.}
      \label{eccv performance comparison}
      \begin{center}
        \begin{tabular}{l|c}
          \hline
          \textbf{Method} &  \textbf{PC} \\ \hline\hline
          Eigenface & 0.180 \\
          Single Layer Model & 0.417 \\
          Two Layer Model & 0.438 \\
          Multiscale Model \cite{gray2010predicting} & 0.458 \\
          Auto Encoder \cite{wang2014attractive} & 0.437 \\
          \textbf{{Ours}} & \textbf{{0.468}} \\ \hline
        \end{tabular}
        \begin{tablenotes}
          \footnotesize
          Performance comparison on ECCV HotOrNot dataset
          \cite{gray2010predicting}.
          Pearson Correlation (PC) is used for evaluating performance.
          Our method achieves the best result on this dataset as mentioned in
          \cite{gray2010predicting}.
        \end{tablenotes}
      \end{center}\label{tb:PC}
    \end{table}

\section{Conclusion}

  In this paper, we propose a method which extracts rich deep facial features
  through knowledge adaptation, and then trains Bayesian ridge regression
  algorithm for face beauty prediction. Despite that the VGG model is pretrained for
  a totally different task, it also captures more descriptive information than
  conventional hand-crafted features, and even outperforms many deep
  learning-based methods in our facial beauty prediction task,
  which shows the great generality of deep features in transfer learning.
  With our feature fusion strategy, our method outperforms other methods and achieves
  the state-of-the-art performance on ECCV HotOrNot dataset~\cite{gray2010predicting}
  without face alignment and comparable performance on SCUT-FBP dataset~\cite{xie2015scut}.
  In our future work, we plan to explore 3D face alignment  and
  novel networks architecture for extracting more descriptive features.


\section*{Acknowledgment}

  This work was primarily supported by Foundation Research Funds for the
  Central Universities (Program No.2662017JC049) and State Scholarship
  Fund (NO.261606765054).



%


\bibliographystyle{IEEEtran}
\bibliography{references}

\end{document}